\documentclass{article}
\usepackage{spconf,amsmath,graphicx,hyperref,url,times,booktabs, tabularx}

\usepackage{color-edits}
\addauthor{leo}{black}

\title{An Evaluation Framework for Structured Audio Captions Validated by Controlled Perturbations}
\name{Liang-Yuan Wu$^{1}$,
      Sripathi Sridhar$^{2}$,
      Mark Cartwright$^{2}$,
      Magdalena Fuentes$^{1}$}
\address{$^{1}$New York University, New York, USA \;
$^{2}$New Jersey Institute of Technology, New Jersey, USA
}
\begin{document}
\ninept
\maketitle
\begin{abstract}

Recent advances in automated audio captioning (AAC) are driving a shift from monolithic sentences toward structured formats that disentangle acoustic and semantic properties, \leoedit{such as timestamped captions for different sound events}. Such representations can support faceted sound search for creators and richer access to auditory information for Deaf and Hard of Hearing people. Yet, it remains unclear how to meaningfully evaluate these hybrid, structured captions. We propose an evaluation framework for structured audio descriptions, spanning five complementary axes: tag sets, descriptions, reasoning, numeric measurements, and spectral profiles. The framework combines large language model (LLM) judges for semantic fields with deterministic metrics for temporal and acoustic attributes. To validate these metrics, we introduce controlled perturbations that apply typed, graded changes to ground-truth annotations. Results show that the proposed metrics remain robust to meaning-preserving paraphrases while responding to genuine semantic and acoustic corruptions, \leoedit{enabling more reliable evaluation of structured captions.}

\end{abstract}

\section{Introduction}
\label{sec:intro}
Recent advancements in automated audio captioning (AAC) have driven a shift from monolithic sentence descriptions~\cite{drossos2017automated, drossos2020clotho,mei2024wavcaps} toward structured formats~\cite{sridhar2026audiocards, niu2026acavcaps, kumar2026tac}. For instance, rather than generating a single, flat sentence, models explicitly isolate distinct acoustic properties such as the sound source, related actions, causative factors, loudness, and duration. This is valuable in creative workflows, where sound search involves relevant facets~\cite{sridhar2026audiocards,weck2024language}. It could also enhance accessibility for accessing useful auditory information for Deaf or Hard-of-Hearing (DHH) people~\cite{wu2025soundnarratives}, as well as serve as an intermediate representation for user-adaptable captioning~\cite{huang2025captune}.

However, evaluating these structured audio descriptions remains a significant challenge~\cite{mei2022automated}. Existing evaluation metrics~\cite{papineni2002bleu,lin2004rouge,banerjee2005meteor} are primarily designed to assess monolithic caption quality through exact n-gram matching. While recent semantic and embedding-based metrics~\cite{zhou2022can, elizalde2023clap} improve robustness to paraphrasing, they still assume a homogeneous textual output. Besides few public datasets jointly combining semantic descriptions with temporal structure and quantitative acoustic annotations~\cite{xie2025audiotime}, comprehensive evaluation of such heterogeneous outputs remains unexplored~\cite{luong2025graph}.

\begin{figure*}[t]
\centering
\includegraphics[width=\textwidth]{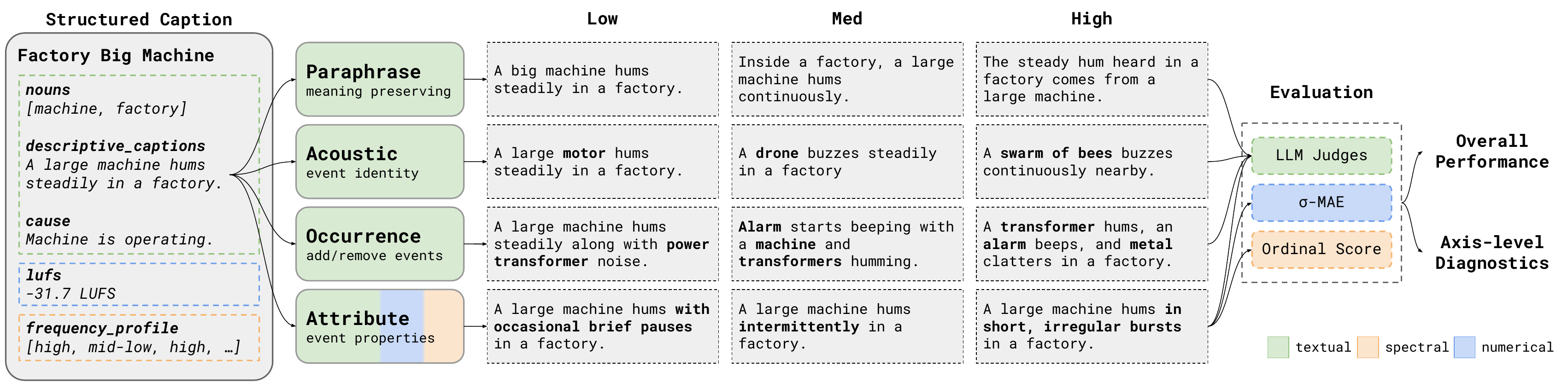}
\caption{\leoreplace{Our evaluation framework. Each AudioCards structured caption is evaluated along five orthogonal axes scored by a purpose-built metric in $[0,1]$. The framework is validated by a controlled perturbation study (four perturbation types $\times$ three severities).}{Overview of the evaluation framework and its perturbation audit. Four perturbation types at three severity levels produce synthetic outputs from 15-field structured captions, which are scored against their references using field-specific metrics. Attribute perturbations can affect all fields and thus connect to all metric branches.}}
\label{fig:overview}
\end{figure*}

To bridge this gap, we propose a comprehensive evaluation framework (Figure \ref{fig:overview}) designed to assess structured audio descriptions across multiple dimensions. We categorize the evaluation space into five distinct axes: tag-sets, free-text descriptions, logical reasoning, numeric measurements, and spectral profiles. To capture semantic nuance without strictly penalizing valid rephrasing, we leverage an LLM-as-a-judge~\cite{zheng2023judging} for the textual and reasoning fields. Conversely, we employ deterministic computational metrics~\cite{kim2018crepe, mesaros2016metrics, caba2015activitynet} to precisely evaluate numerical and acoustic measurements. Each axis is evaluated by a field-specific metric that is normalized to a commensurable $[0,1]$ scale, facilitating seamless aggregation into a unified score. The benchmark score supports overall model comparison, whereas the axis and field scores provide a diagnostic breakdown of specific capabilities.

As an initial step toward validating the framework, we conduct a controlled perturbation study using \textit{AudioCards}~\cite{sridhar2026audiocards}, a recently released dataset providing multi-field structured information tailored to sound design needs. Its annotations are organized according to the Universal Category System~\cite{ucs}, which we complement with signal-derived acoustic measurements. We establish that an ideal metric must remain robust to meaning-preserving paraphrases while maintaining acute sensitivity to genuine semantic and acoustic errors. To rigorously audit these metrics, we introduce a synthetic perturbation testing protocol~\cite{ribeiro2020beyond} that injects \leoedit{different types of errors into the dataset, stratified into severity levels according to their expected perceptual impact}. Using this protocol, we demonstrate our framework's ability to distinguish between valid linguistic variation and true model failure. Finally, we analyze the computational and diagnostic trade-offs between LLM judges and traditional NLP metrics, providing concrete recommendations for their optimal deployment.

Our main contributions are threefold: (i) We introduce a unified evaluation framework capable of assessing the diverse, multi-dimensional aspects of structured audio descriptions; (ii) We establish a systematic perturbation testing protocol designed to audit metric reliability across varying error types and severities reflecting human perceptions; and (iii) We present a comprehensive empirical analysis comparing the efficacy of LLM judges against traditional metrics to guide future evaluation strategies.

\section{Methods}
\label{sec:task}
\subsection{Dataset}
Unlike standard audio captioning datasets, AudioCards~\cite{sridhar2026audiocards} provides professionally curated sound effects (SFX) annotated with multi-field, structured attributes organized under the Universal Category System (UCS). This structured format is essential for our diagnostic benchmark, as it decomposes a rich acoustic structure into discrete, independently evaluable properties.

While AudioCards provides 11 semantic and descriptive annotation fields, it does not include signal-derived low-level acoustic reference measurements, \leoedit{such as loudness and spectral characteristics, which capture perceptually salient information not explicitly represented in captions. Such acoustic information could inform personalized captioning; for example, a frequency profile may help prioritize content in bands where a listener has reduced hearing sensitivity.} To provide these missing acoustic references, we augment each clip with four automatically computed acoustic measurements: integrated loudness (LUFS); median fundamental frequency (343 clips with pitched content, while the remaining 156 contain no dominant pitch); active onset and offset bounds (active event segments); and a ten-band frequency profile. We treat the resulting \textbf{15 fields} as our evaluation targets. We will release this augmented AudioCards dataset to encourage further research on structured audio captioning. 

\subsection{Tasks}

\leoedit{To evaluate structured captions, the framework must distinguish valid linguistic variation from substantive errors. Inspired by behavioral testing methodologies in natural language processing~\cite{ribeiro2020beyond} and recent perturbation-based evaluations in audio captioning~\cite{dixit2025mace, yermekovatesting}, we audit this ability through a controlled perturbation study. By isolating error modes, we test whether each evaluation axis responds selectively to its corresponding perturbations and tracks their intended severity.}

\leoedit{We frame the metric audit as a structured-output diagnosis task. For each reference annotation, we generate \textit{synthetic model outputs} by applying controlled perturbations at predefined severity levels. Unlike outputs from actual models, these outputs' error types and degrees of correctness are known by construction, with each perturbation designed to represent a perceptually motivated level of error.} Because these 15 fields encompass fundamentally different data formats, we organize them into five distinct data types, each requiring a tailored evaluation protocol: \textit{Tag-set}, \textit{Description}, \textit{Reasoning}, \textit{Numeric}, and \textit{Spectral} (see Table~\ref{tab:columns}).

\begin{table}[t]
\small
\centering
\caption{The 15 evaluation fields, grouped by data type each family uses. $\bar{w}$: mean word count; $|\bar{T}|$: mean tag cardinality.}
\label{tab:columns}
\setlength{\tabcolsep}{4pt}
\footnotesize
\begin{tabular}{@{}ll@{}}
    \toprule
    \textbf{Field} & \textbf{Detail} \\
    \midrule
    \multicolumn{2}{@{}l}{\textit{Tag-set} (5)} \\
    \quad \texttt{nouns}                & $|T|=4.1$ \\
    \quad \texttt{verbs}                & $|T|=2.7$ \\
    \quad \texttt{noun\_verb\_pairs}    & $|T|=2.9$ \\
    \quad \texttt{adjectives}           & $|T|=4.3$ \\
    \quad \texttt{complementary\_sounds} & $|T|=4.4$ \\
    \midrule
    \multicolumn{2}{@{}l}{\textit{Description} (3) } \\
    \quad \texttt{caption\_in\_3\_words}          & $\bar{w}=3.2$ \\
    \quad \texttt{caption\_in\_7\_words\_or\_less} & $\bar{w}=5.6$ \\
    \quad \texttt{descriptive\_caption}           & $\bar{w}=19.0$ \\
    \midrule
    \multicolumn{2}{@{}l}{\textit{Reasoning} (3)} \\
    \quad \texttt{cause}                  & free text \\
    \quad \texttt{effect}                 & free text \\
    \quad \texttt{example\_visual\_context} & free text \\
    \midrule
    \multicolumn{2}{@{}l}{\textit{Numeric} (3)} \\
    \quad \texttt{lufs}            & $-51.5$ to $-7.7$ dBFS \\
    \quad \texttt{median\_f0}      & 72.4--132.4 semitones \\
    \quad \texttt{active\_event} & list of (onset, offset) spans\\
    \midrule
    \multicolumn{2}{@{}l}{\textit{Spectral} (1)} \\
    \quad \texttt{frequency\_profile} & 10 bands, 4 ordinal levels \\
    \bottomrule
\end{tabular}
\end{table}

\section{Controlled Perturbation Study}
\label{sec:perturbation}

\subsection{Perturbation Types and Severity}
We structure our perturbation space as a two-dimensional matrix of type and severity. \leoedit{\emph{Paraphrase perturbation} rewrites all textual field families while preserving meaning. \emph{Acoustic perturbation} substitutes the reference sound source with a different source, updating source-related tags, descriptions, and reasoning. \emph{Occurrence perturbation} inserts unsupported events or removes reference events in textual fields, leaving numerical and spectral fields unchanged. \emph{Attribute perturbation} changes perceptual properties such as loudness, pitch, temporal pattern, or timbre, updating the textual, numerical, and spectral fields that encode them.}

\leoedit{Each perturbation type has three severity levels: low, medium, and high (see Table \ref{tab:severity}), whose interpretation depends on the type of change. For paraphrases, severity denotes the extent of rewording while meaning remains unchanged. For acoustic, occurrence, and attribute perturbations, severity is operationalized by field family (Table~\ref{tab:severity}): the number of edited tag-set items, the extent of semantic deviation in descriptions and reasoning, the magnitude of scalar numerical changes, and the number and ordinal shift of modified spectral bands.}


\begin{table}[t]
\centering
\caption{Intended severity of perturbations by field family.}
\label{tab:severity}
\small
\setlength{\tabcolsep}{3pt}
\begin{tabularx}{\columnwidth}{@{}
  >{\raggedright\arraybackslash}p{0.36\columnwidth}
  *{3}{>{\centering\arraybackslash}X}
@{}}
\toprule
\textbf{Field family} & \textbf{Low} & \textbf{Med.} & \textbf{High} \\
\midrule
Tag-set & 1 item & 2--3 items & all items \\
Description, Reasoning & minor & moderate & major \\
Numerical & $\pm0.25\sigma$ & $\pm1.5\sigma$ & $\pm3\sigma$ \\
Spectral & $(2,\pm1)$ & $(5,\pm2)$ & $(10,\pm2)$ \\
\bottomrule
\end{tabularx}
\end{table}

\subsection{Implementation}

\textbf{Generation.} Text perturbations are produced by Meta-Llama-3-70B-Instruct~\cite{llama3modelcard} with calibrated prompts and greedy decoding for reproducibility. As manual quality control, we inspected around $5\%$ perturbed structured captions, spanning different error types and severity levels, and refined the prompts when outputs failed to preserve the intended meaning or match the specified corruption. Numeric and frequency perturbations are exact arithmetic shifts, so the injected error is known precisely. 

\textbf{Baselines.} We compare each proposed metric with existing evaluation metrics suited to each field type. For the caption and reasoning families we use the lexical-overlap scores BLEU-1~\cite{papineni2002bleu}, ROUGE-L~\cite{lin2004rouge}, and METEOR~\cite{banerjee2005meteor} together with the embedding-based FENSE~\cite{zhou2022can}. For tag-sets we use exact set-F1 and CLAP text-embedding soft-F1 (a predicted--reference cosine matrix greedily matched above a threshold, reported at $\tau\in\{0.5,0.7\}$)~\cite{wu2023large}. For the numeric fields the baseline is the raw MAE in interpretable units (dB, semitones, seconds) prior to $\sigma$-normalization, and for the spectral field it is per-band exact-match accuracy, which counts any label change as a full miss.

\section{Field-Specific Evaluation Metrics}
\label{sec:metrics}
We use one evaluation protocol for each of the five field families. Each returns a score in $[0,1]$; we aggregate these scores to summarize overall performance and report axis-level results to identify specific strengths and failure modes We report each of these five metric families as described in Figure~\ref{fig:overview}.

\textbf{Tag-set Fields.}
We evaluate tag-sets (e.g., nouns, verbs) using a single-pass \emph{LLM judge}. The LLM is prompted with the entire set of nouns, verbs, and it is asked to compare to the reference set and come up with a score in $[0,1]$. By reasoning over both sets as a whole, the judge is able to match valid synonyms and near-synonyms (e.g.\ sound\,$\leftrightarrow$\,noise, metallic\,$\leftrightarrow$\,industrial) while jointly penalizing \emph{misses} (reference tags with no match) and \emph{hallucinations} (predicted tags matching nothing). To ground the evaluation, the judge is provided with contexts of what field it is evaluating (e.g., ``sound-source nouns'' or ``acoustic-quality adjectives'').

\textbf{Description Fields.}
\leoedit{Description fields contain free-text captions describing the audio, with some fields imposing explicit word-count constraints.}
We evaluate descriptions using an LLM judge that rates correctness, relevance, completeness, and clarity~\cite{kumar2026mmau}; their mean gives the semantic score.For length-constrained captions, we multiply this mean by $\lambda=\max(0,1-v/k)$, where $v$ is the word-count violation (the deviation from the required length) and $k$ is the specified word budget. Unconstrained descriptions use $\lambda=1$.

\textbf{Reasoning Fields.}
\leoedit{Reasoning fields contain free-text contextual inferences about the audio, including its likely cause, its perceptual effect on a listener, and a plausible visual scene that could reasonably produce or accompany the sound. We leverage the LLM judge to assess whether each response is coherent, plausible, and consistent with the reference rather than whether it uses the same phrasing.} The judge applies the same four-criteria rubric (correctness, relevance, completeness, and clarity) used for descriptions, but is guided by reasoning-specific field hints. \leoedit{The resulting score should reflect the faithfulness and completeness of the explanation rather than its lexical overlap with the reference.}

\textbf{Numerical Fields.}
\leoedit{Numerical fields comprise signal-derived scalar measurements and temporal event intervals. Because their expected responses to controlled perturbations follow directly from the metric definitions, we include these experiments primarily for completeness and to verify the calibration and integration of the numerical scores within the overall framework.}

\textit{Scalar fields (\textit{lufs}, \textit{median\_f0}).} To evaluate scalar attributes, we compute the raw Mean Absolute Error (MAE) between the prediction $\hat{y}$ and the ground truth $y$. For \textit{lufs}, this error is calculated directly in decibels (dB). For \textit{median\_f0}, we first map the frequencies to a pitch-linear scale ($12\log_2 f$) before computing the error, ignoring clips without pitched content.

\textit{Normalization:} To project these unconstrained errors onto a unified $[0,1]$  axis, we apply a $\sigma$-normalized transformation:

\begin{equation}
    \label{eq:norm_mae}
    s = \mathrm{clip}\!\left(1 - \frac{|\hat{y} - y|}{4\sigma},\ 0,\ 1\right)
\end{equation}

Here, $\sigma$ represents the dataset's robust dispersion, computed once over the full dataset using the median absolute deviation ($\sigma=6.86$ for \textit{lufs}, $\sigma=14.97$ for \textit{median\_f0}). Assuming the attribute values are approximately Gaussian-distributed, we set the normalization range to $4\sigma$, corresponding to the width of the interval $\mu\pm2\sigma$ that contains approximately $95\%$ of the distribution.

\textit{Temporal activity (\textit{active\_event}).} We evaluate the timing of active event lists by treating them as multi-segment temporal localization tasks. We greedily form one-to-one matches between the predicted set $\hat{E}$ and the reference set $E$. We measure performance using two metrics that natively scale to $[0,1]$.

\textit{Collar-based F1:} Following the standard audio event detection~\cite{mesaros2016metrics}, a predicted segment matches a reference if its onset and offset boundaries both fall within a specific temporal tolerance ($\tau_{\text{on}}=\tau_{\text{off}}=0.2$s, with the offset tolerance widening dynamically for longer references). We pool these matches across the dataset to report the standard micro and macro F1 scores.

\textit{Count-aware mIoU:} Instead of discrete boundary tolerances, segments are matched by their continuous intersection-over-union (IoU)~\cite{caba2015activitynet}. To penalize models for hallucinating extra segments or missing ground-truth events, we normalize the total matched IoU mass by the maximum number of segments present in either the prediction or the reference:

\begin{equation}
    \label{eq:miou}
    \text{mIoU} \;=\; \frac{1}{N}\sum_{c=1}^{N}
    \frac{\sum_{(\hat{\imath},\jmath)\in M_c}\mathrm{IoU}(\hat{\imath},\jmath)}
         {\max\!\big(|\hat{E}_c|,|E_c|\big)},
\end{equation}

\textbf{Spectral Fields.}
\leoedit{The spectral field captures an audio clip's frequency profile.  We represent the profile as} a JSON object mapping ten octave-band center frequencies (125–10000 Hz) to ordinal labels $\{$low, medium-low, medium-high, high$\}$. We map these to integers ${0,1,2,3}$ and compute:

\begin{equation}
    \label{eq:f_profile}
    s = 1 - \frac{1}{10}\sum_{b} \frac{|\hat{o}_b - o_b|}{3}
\end{equation}

yielding 1 for an exact match and 0 for maximum disagreement on every band. This ordinal score degrades smoothly as the spectral shape is corrupted.

\section{Results}
\label{sec:results}
\textbf{Setup.} We use Qwen2.5-7B-Instruct~\cite{qwen2.5} as the primary LLM judge and compare it with baseline metrics on controlled perturbations of ground-truth (GT) AudioCards annotations. Perturbations are applied at low, medium, and high intensities. Scalar errors are mapped to $[0,1]$ using field-specific normalization, while temporal and spectral fields use deterministic metrics.

\begin{figure}[h]
\centering
\includegraphics[width=\linewidth]{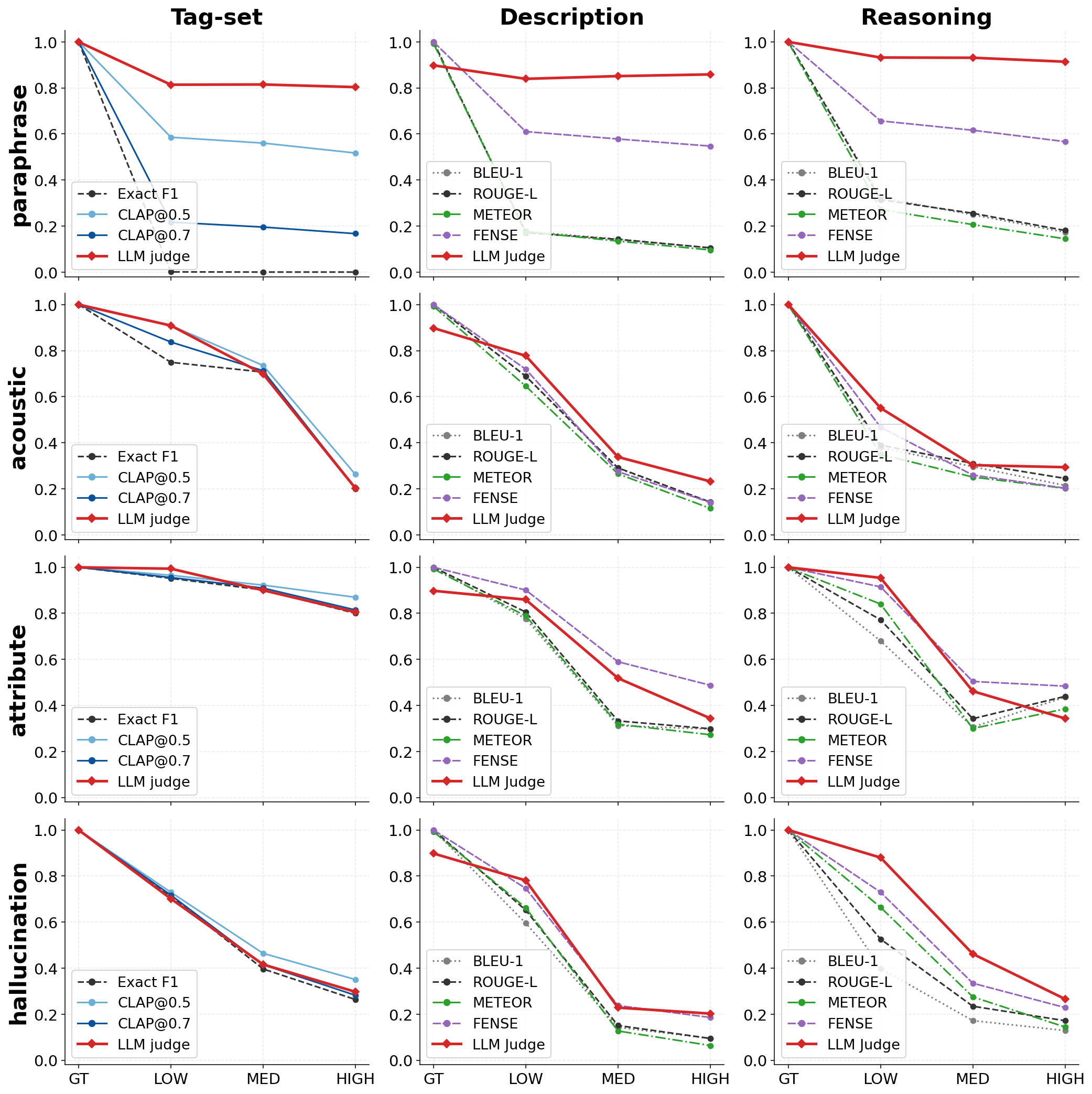}
\caption{Text-metric responses under four perturbation types.}
\label{fig:llm_results}
\end{figure}

\textbf{Evaluation results.} For text fields, an ideal evaluation metric should isolate genuine errors without penalizing valid linguistic variations. In Figure~\ref{fig:llm_results}, our results demonstrate that the LLM judge outperforms traditional metrics (a more flat trending curve) in handling meaning-preserving paraphrases. In contrast, under true semantic corruptions (acoustic, attribute, and occurence), the LLM judge performs on par with traditional metrics, exhibiting strict monotonic degradation that accurately tracks error severity.

\begin{figure}[h]
\centering
\includegraphics[width=\linewidth]{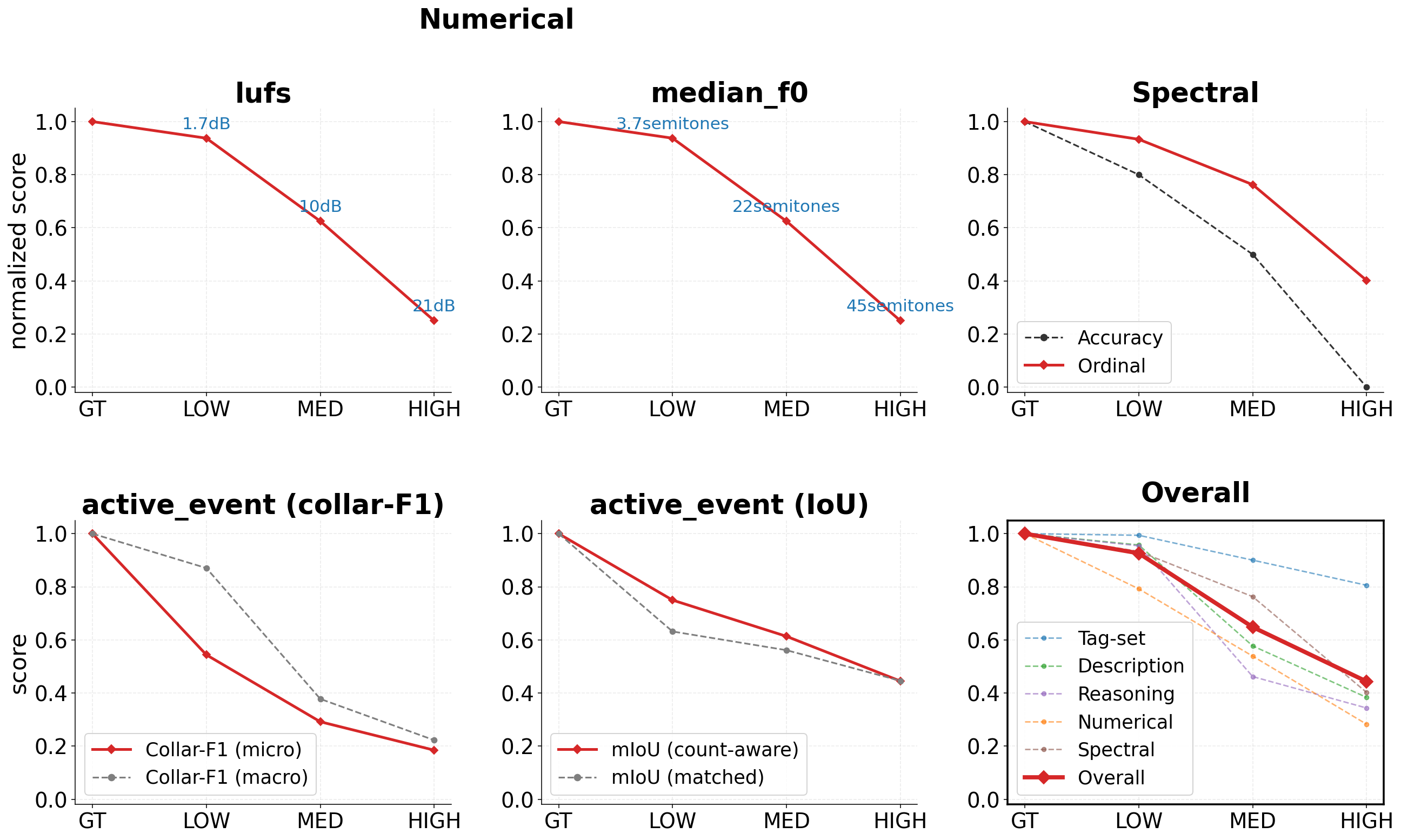}
\caption{Responses for numerical/spectral and aggregate metrics to increasing perturbation severity.}
\label{fig:numeric_results}
\end{figure}

As expected, all numerical scores decrease as perturbation magnitude increases (Figure~\ref{fig:numeric_results}). The ordinal spectral score preserves graded label differences, whereas exact match treats all mismatches equally. These results confirm that each metric responds to controlled perturbations as specified by its formulation. 
\leoedit{The overall score averages the five field-family scores to summarize structured captions quality. The monotonic decline with perturbation severity tracks the expected ordering of the synthetic outputs. This establishes internal consistency under controlled perturbations, although further validation against human judgments is needed to determine how well the aggregate reflects caption quality in practice.}

\textbf{Judge model analysis.} To assess whether the results depend on the choice of LLM judge, we applied the same evaluation prompts to three additional instruction-tuned models: Qwen2.5-0.5B-Instruct~\cite{qwen2}, Qwen3-4B-Instruct~\cite{qwen3technicalreport}, and a similar sized counterpart, Meta-Llama-3-8B~\cite{llama3modelcard}. Their Pearson correlations with the Qwen2.5-7B scores are $r=0.575$, $r=0.983$, and $r=0.971$, respectively. Manual inspection revealed that the smallest model, Qwen2.5-0.5B-Instruct, sometimes failed to follow the evaluation instructions, yielding outputs that were irrelevant to the task.

To assess inference cost, we also run the complete evaluation pipeline over 100 AudioCards examples on one NVIDIA L40S GPU (Table~\ref{tab:inference}). Under the current prompting and decoding setup, Qwen2.5-0.5B is the fastest and uses the least memory, but produces valid JSON for only $23\%$ of outputs; the other three models achieve $100\%$ formatting success on this sample. Among these models, Qwen3-4B uses the least peak memory (10.44\,GB), whereas Qwen2.5-7B is slightly faster. Qwen2.5-7B also uses the least energy among these models, at 415 joules (J) per structured caption, compared with 422 J for Qwen3-4B and 556 J for Llama-3-8B. Ultimately, we conclude that our framework is highly portable across moderate-sized instruction-tuned models, whereas tiny LLMs currently lack the formatting and reasoning stability required for automated evaluation.

\begin{table}[t]
\centering
\caption{Inference cost of judging structured caption (S. caption).}
\label{tab:inference}
\footnotesize
\begin{tabular}{lcccc}
    \toprule
    \textbf{Model} & \textbf{s/S. caption} & \textbf{J/S. caption} & \textbf{Peak mem (GB)} & \textbf{Valid} \\
    \midrule
    Qwen2.5-0.5B & 0.86 & 175 & 1.55 & 23\% \\
    Qwen2.5-7B   & 1.31 & 415 & 17.35 & 100\% \\
    Qwen3-4B     & 1.41 & 422 & 10.44 & 100\% \\
    Llama-3-8B   & 1.74 & 556 & 18.73 & 100\% \\
    \bottomrule
\end{tabular}
\end{table}

\textbf{The cost-sensitivity trade-off.} Selecting an evaluation protocol requires balancing computational cost, diagnostic sensitivity, and target-model quality. \leoedit{Efficient traditional metrics, such as BLEU and cosine similarity, may suffice for systems that produce simple, coarse-grained captions. For more capable systems that generate richer and more linguistically diverse descriptions, however, these metrics may penalize valid paraphrases or miss subtle semantic errors. LLM judges incur greater computational cost but better capture these distinctions, making them more suitable for fine-grained evaluation of advanced audio-language models.}

\section{Conclusion}
\label{sec:conclusion}
We introduced a field-specific evaluation framework and systematic perturbation protocol to address the limitations of traditional NLP metrics in structured audio captioning. Our findings demonstrate that an LLM-as-a-judge successfully balances semantic flexibility with diagnostic rigor, effectively capturing paraphrasing nuances while penalizing genuine acoustic and textual corruptions. These experiments constitute an initial behavioral validation. Moving forward, we will investigate the generalization of this evaluation framework across more diverse audio types beyond sound effects, and benchmark its performance directly on real-world outputs from emerging audio-language models to further solidify its utility in practical deployment. \leoreplace{Ultimately, we envision this evaluation standard not just as a diagnostic tool, but as a foundational metric that could be integrated directly into training loops to actively guide the development of future audio-language models.}{As audio-language models and their associated tasks become complex, systematic audits of evaluation metrics will be essential to ensuring reliable and meaningful assessment.}

\clearpage
\bibliographystyle{IEEEbib}
\bibliography{strings,refs}

\end{document}